\documentclass[pdflatex,sn-mathphys-num]{sn-jnl}

\usepackage{graphicx}%
\usepackage{multirow}%
\usepackage{amsmath,amssymb,amsfonts}%
\usepackage{amsthm}%
\usepackage{mathrsfs}%
\usepackage[title]{appendix}%
\usepackage{xcolor}%
\usepackage{textcomp}%
\usepackage{manyfoot}%
\usepackage{booktabs}%
\usepackage{algorithm}%
\usepackage{algorithmicx}%
\usepackage{algpseudocode}%
\usepackage{listings}%

\theoremstyle{thmstyleone}%
\theoremstyle{thmstyletwo}%

\theoremstyle{thmstylethree}%

\begin{document}

\title[Article Title]{HarmoniDPO: Video-guided Audio Generation via Preference-Optimized Diffusion}


\author[1]{\fnm{Wenshuo} \sur{Peng}}\email{gin2pws@gmail.com}

\author*[2]{\fnm{Kaipeng} \sur{Zhang}}\email{
kpzhang@ut-vision.org}


\affil[1]{\orgdiv{Electornic Engineering Department}, \orgname{Tsinghua University}, \orgaddress{\street{Haidian District}, \city{Beijing}, \postcode{100084} \country{China}}}

\affil*[2]{\orgdiv{OpenGVlab}, \orgname{Shanghai AI Laboratory}, \orgaddress{\street{Xuhui District}, \city{Shanghai}, \postcode{200030}, \country{China}}}



\abstract{Video-to-audio (V2A) generation faces significant challenges in achieving precise temporal synchronization and high perceptual quality due to the complex, ambiguous relationship between visual and auditory cues. Existing methods typically compress video inputs into single feature representations, leading to significant loss of temporal dynamics and fine-grained visual information. These approaches also rely on reconstruction-based training objectives that poorly correlate with human perceptual judgments of audio quality and appropriateness. We propose HarmoniDPO, a novel framework that integrates preference-based optimization into diffusion-based V2A generation to address these limitations. (1) Our approach leverages a dual video representation: combining global context with frame-wise features to preserve temporal dynamics and semantic detail. (2) Inspired by reinforcement learning from human feedback (RLHF), HarmoniDPO employs online Direct Preference Optimization (online-DPO) to fine-tune a diffusion-based V2A model from preference judgments, enhancing perceptual quality and alignment. (3) Additionally, we introduce Dual-scale Diffusion Search (DDS), a test time scaling algorithm that adaptively optimizes output fidelity during inference. Experiments demonstrate that HarmoniDPO outperforms state-of-the-art methods in audio-video synchronization and subjective audio quality, offering a robust solution for generating realistic, human-preferred audio from video.  }




\maketitle

\section{Introduction}\label{sec1}
Recent advancements in cross-modal learning have led to significant progress in audio generation. State-of-the-art models like AudioGen~\citep{kreuk2022audiogen}, Make-An-Audio~\cite{huang2023make}, and AudioLDM~\cite{liu2023audioldm} have demonstrated remarkable capabilities in synthesizing audio from text descriptions. As text-to-video generation~\cite{khachatryan2023text2video, ge2023preserve} continues to show promising applications, there is increasing interest in tackling the inverse problem: generating audio from video.

Video-to-audio generation presents significantly greater challenges than text-to-audio generation, primarily due to the following reasons: (1) Multimodal Alignment Requirements: The generated audio must synchronize with both the semantic content and temporal dynamics of the video. (2) Ambiguity in Audio-Visual Mapping and Perceptual Quality: The relationship between visual events and corresponding audio is inherently complex and often ambiguous. Visual content alone frequently under specifies the full range of desired audio characteristics, encompassing not just semantic alignment but also crucial aspects of perceptual quality, realism, and overall appropriateness. Determining the optimal audio solely from video cues is challenging, as multiple soundscapes might fit the visuals yet vary significantly in perceived quality without further optimization criteria.

\begin{figure}[t]
    \centering
    \includegraphics[width=1\textwidth]{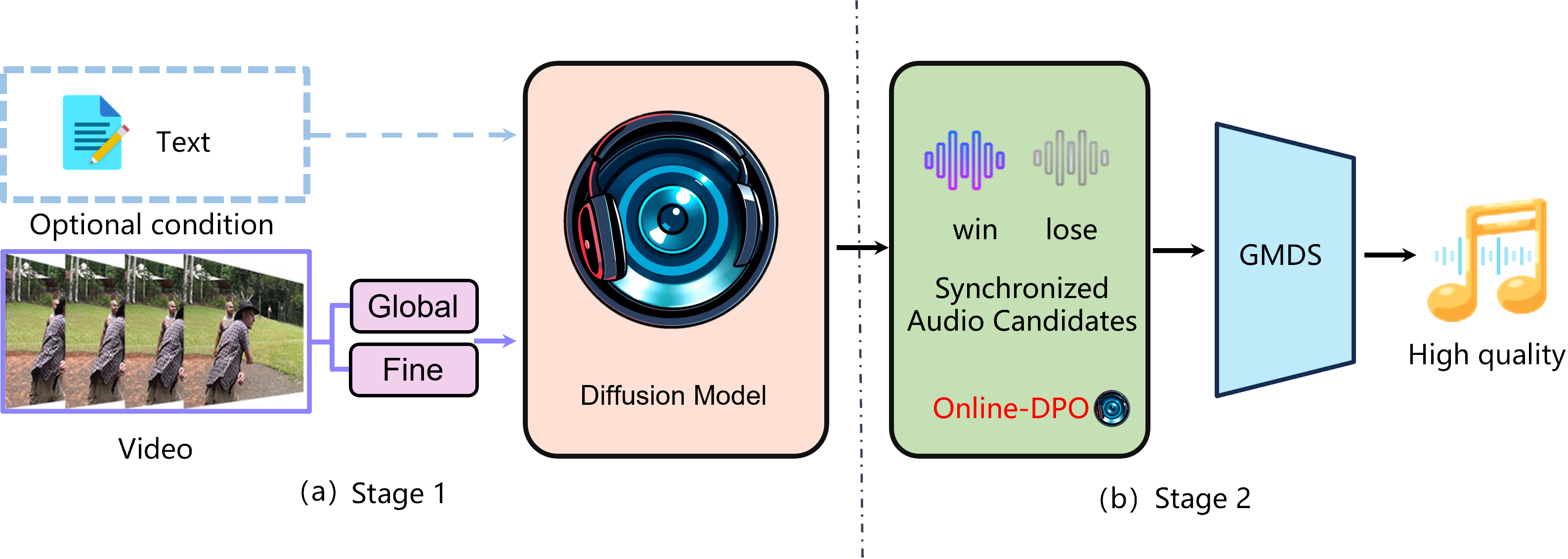}
    \caption{Overview of the HarmoniDPO framework. (a) Stage 1: The diffusion model processes dual video inputs (global context features and frame-wise local features) with optional text conditioning to generate initial audio. (b) Stage 2: Online Direct Preference Optimization fine-tunes the model using pairwise audio comparisons to align outputs with preferences.}
    \label{fig:1}
\end{figure}

Several methods have been proposed for video-to-audio generation, including diffusion-based architectures \cite{zhang2024foleycrafter}, contrastive learning approaches \cite{luo2024diff}, and text-mediated techniques \cite{wang2024v2a}. While these methods have shown progress, a critical limitation persists: ensuring the generated audio achieves high perceptual quality and aligns robustly with subjective human preferences regarding realism, style, and appropriateness for the given video context. The core issue is that standard training paradigms, often relying on objectives like L1 or L2 reconstruction loss against a single ground truth audio, are fundamentally insufficient. Such objectives act as poor proxies for complex human auditory perception; they struggle to capture the nuances that differentiate acceptable audio from truly high-quality, compelling audio, especially given the inherent ambiguity where multiple sounds could plausibly match the visuals. This reliance on simplistic loss functions fails to provide a strong alignment signal or reward for optimizing towards perceptually superior outcomes preferred by humans. Consequently, models trained this way may generate technically synchronized audio that lacks realism or fails to meet subjective quality standards. 
Beyond the fundamental limitations of reconstruction-based training, an even deeper issue emerges from the very nature of video-to-audio generation. The field suffers not only from data scarcity but also from an intrinsic mismatch between available training signals and desired model behavior. While architectural innovations have made progress in addressing temporal synchronization challenges, they fail to resolve the core disconnect between what standard training optimizes for (reconstruction of potentially noisy reference audio) and what ultimately matters (human perception of quality and appropriateness). This misalignment becomes particularly problematic given the one-to-many nature of video-to-audio mapping - where multiple plausible soundtracks could accompany the same visuals, each varying in subjective quality and stylistic preference. The critical bottleneck, therefore, lies not in the model's capacity to reconstruct audio, but in its ability to discern and generate human-preferred outputs from this space of valid possibilities. This realization compels us to move beyond conventional paradigms and adopt preference-driven approaches that can directly optimize for perceptual quality, leveraging comparative judgments to guide the model toward solutions that better align with human evaluation criteria. 

To overcome the limitations of current approaches, we introduce HarmoniDPO, an effective diffusion-based V2A model with a novel framework that, for the first time, applies preference-based optimization to video-to-audio (V2A) generation. Inspired by Reinforcement Learning from Human Feedback (RLHF), our method employs online Direct Preference Optimization (online-DPO) through a two-stage process. 
(1) First, we develop an effective diffusion-based V2A model (illustrated in Figure~\ref{fig:1}-Stage1), which is built upon a U-Net diffusion architecture. Innovatively, HarmoniDPO processes dual video inputs: a global feature capturing overall context and fine-grained frame-wise features to preserve detail and facilitate precise temporal alignment, mitigating information loss common in single-feature representations. HarmoniDPO also supports optional text conditioning. 
(2) In the second stage, we fine-tune this pre-trained HarmoniDPO latent diffusion model (LDM) using our online-DPO strategy. This involves learning directly from preference comparisons between pairs of generated audio samples. The model is iteratively refined based on a learned reward function derived from these preferences, implicitly capturing human judgments of audio quality and video-audio congruence. This preference-based fine-tuning enables HarmoniDPO to generate audio that is perceptually superior and better aligned with desired characteristics (e.g., temporal synchrony, style consistency) beyond what standard training achieves, allowing it to effectively explore the output space and converge towards solutions reflecting human preferences.

Besides the model architecture design, we also address these challenges at the inference stage with a test-time scaling strategy inspired by advancements in the text-to-image domain \cite{ma2025inference}. Specifically, we propose a Dual-scale Diffusion Search (DDS) algorithm, which employs a dual-scale sampling mechanism to explore the latent space adaptively. This approach balances local exploitation and global exploration by combining small and large step sizes, ensuring more efficient and effective optimization during inference. We conduct experiments to demonstrate the superiority of our method, and our contributions can be summarized as follows:
\begin{itemize}
\item We identify that maintaining both global context and temporal dynamics is crucial for video-to-audio generation. Based on this insight, we propose a simple yet effective diffusion-based V2A model that uses a complementary representation of global video features and frame-wise image features, significantly improving audio-video synchronization quality. Moreover, it accepts optional text inputs for further quality improvement and precise control.
\item We identify that ensuring the generated audio achieves high perceptual quality and aligns robustly with subjective human preferences is a critical limitation. Hence, we pioneer the application of preference-based optimization to the V2A generation task, proposing an online Direct Preference Optimization (online-DPO) framework inspired by RLHF. This significantly enhances perceptual audio quality and alignment by fine-tuning based on human preferences.
\item We develop a test-time scaling method specifically designed for video-to-audio generation, which optimizes the model's output quality during inference. This method ensures high fidelity and temporal alignment of the generated audio with the input video.
\item Experiments demonstrate that our method outperforms state-of-the-art methods in audio-video synchronization and subjective audio quality, offering a robust solution for generating realistic, human-preferred audio from video.
\end{itemize}

\section{Related Work}
\label{sec:bg }
\subsection{Audio Generation}
\label{sec:bg:audio_gen }

Audio generation has emerged as a prominent research direction in recent years, encompassing both text-to-audio and video-to-audio generation tasks. Contemporary approaches in this field can be systematically categorized into two predominant frameworks: transformer-based autoregressive models and diffusion-based architectures. These frameworks have been applied to various audio generation tasks, including text-conditioned audio synthesis, music generation, and video-synchronized audio generation. 

\textbf{Text-to-Audio Generation.} Text-to-audio generation focuses on synthesizing audio content conditioned on textual descriptions. \textcolor{blue}{Early work in this domain includes Jukebox\cite{dhariwal2020jukebox}, which combined hierarchical VQ-VAE compression with autoregressive Transformers to produce diverse, high-fidelity music conditioned on genre, lyrics, and artist style.} AudioGen~\cite{kreuk2022audiogen}, which pioneered an autoregressive framework for audio synthesis, capable of generating acoustic content from either textual descriptions or audio prompts. Expanding this paradigm, MusicGen~\cite{copet2024simple} implements a sophisticated language modeling approach utilizing efficient token interleaving strategies to enable high-fidelity music generation from textual inputs. \textcolor{blue}{Voicebox\cite{le2023voicebox} represents a breakthrough as the first truly general-purpose speech generation model. By training on 110K hours of diverse audiobook data using a non-autoregressive flow-matching approach, it achieves state-of-the-art performance on zero-shot text-to-speech, speech editing, and cross-lingual synthesis without requiring style labels. MAGNET\cite{ziv2024masked} introduces masked generative modeling with non-autoregressive Transformers, operating on multi-stream audio representations. By iteratively predicting and rescoring masked token spans, it achieves comparable quality to autoregressive baselines while significantly reducing latency.}

Diffusion models have achieve remarkable cross-model generation ability in many works\cite{peng2024t3m,zhang2025v2edit}. In text-to-audio domain, DiffSound~\cite{yang2023diffsound} establishes a comprehensive framework integrating multiple specialized components: a text encoder, a Vector Quantized Variational Autoencoder (VQ-VAE), a token decoder, and a vocoder. DiffSound uses a discretized VQ-VAE for encoding and decoding, combined with a non-autoregressive token decoder based on discrete diffusion. Building on these advancements, AudioLDM~\cite{liu2023audioldm} advances the field by implementing contrastive language-audio pretraining (CLAP) embeddings as latent conditional variables within a VAE framework for audio synthesis. Make-an-Audio~\cite{huang2023make} introduces an innovative approach through a spectrogram autoencoder that predicts self-supervised representations instead of direct waveforms, thereby facilitating enhanced compression efficiency and deeper semantic interpretation. Through the integration of CLAP embeddings~\cite{wu2023large} with high-fidelity diffusion architectures, Make-an-Audio achieves sophisticated language comprehension capabilities alongside superior audio generation quality.
\textcolor{blue}{AudioLDM2\cite{liu2024audioldm} proposes a unified framework that leverages a general audio representation called the ``language of audio" (LOA), derived from the self-supervised AudioMAE model, which employs GPT-2 to translate multimodal inputs into LOA and utilizes a LDM for conditional audio generation.  Stable Audio\cite{evans2024fast} introduces timing embeddings alongside text conditioning, allowing variable-length generation while maintaining musical structure.}

\textbf{Video-to-Audio Generation.} Video-to-audio generation aims to learn joint representations of visual and auditory modalities to produce audio synchronized with visual content. \textcolor{blue}{Early works like Visually Indicated Sounds \cite{owens2016visually} developed a dataset of object interaction videos (hitting, scratching) and proposed a recurrent neural network that maps visual sequences to audio features, which are then converted to waveforms through either exemplar matching or parametric inversion. SpecVQGAN~\cite{SpecVQGAN_Iashin_2021} presents an efficient framework for multi-class, visually guided sound synthesis using a transformer decoder trained to sample from a codebook-based prior. Foley Music\cite{gan2020foley} propose cross-modal music generation frameworks that map body keypoint movements to MIDI representations for synchronized music generation from silent videos. }

\textcolor{blue}{Recent diffusion-based approaches have shown significant progress, with Diff-foley~\cite{luo2024diff} employing latent diffusion models utilizing contrastive audio-visual pre-trained features, while FoleyCrafter~\cite{zhang2024foleycrafter} uses IP-adapter~\cite{ye2023ip} connections to capture video semantics and employs audio event detection for temporal information. Frieren\cite{wang2024frieren} proposes rectified flow-matching frameworks achieving superior quality through straight-trajectory ODE sampling and cross-modal feature fusion.} V2A-Mapper~\cite{wang2024v2a} leverages pre-trained foundation models to bridge multimodal gaps for better open-domain handling.

\textcolor{blue}{Transformer-based methods include MaskVAT\cite{pascual2024masked}, which introduces masked generative transformers combining full-band neural audio codecs with multi-modal conditioning, and V-AURA\cite{viertola2025temporally}, which proposes autoregressive frameworks using discrete audio tokens to avoid spectrogram-based information loss}. \textcolor{blue}{Specialized applications focus on different audio types: Video2Music\cite{kang2024video2music} and Vidmuse\cite{tian2025vidmuse} target music synthesis through affective multimodal transformers and LSTV-modules respectively, while VarietySound\cite{cui2023varietysound} disentangles audio into temporal, timbre, and background components for sound effects generation. TiVA\cite{wang2024tiva} uses unique audio layouts for prediction, Tri-Ergon\cite{li2025tri} achieves precise loudness regulation via LUFS integration, and ReWaS\cite{jeong2025read} introduces energy-based ControlNet adapters for temporal alignment}.


\textcolor{blue}{Current video-to-audio methods, whether reconstruction-based \cite{owens2016visually}, transformer-based \cite{viertola2025temporally}, or diffusion-based \cite{luo2024diff,zhang2024foleycrafter}, share a common limitation: they optimize for reconstruction fidelity against ground-truth audio using standard losses (L1, L2), which poorly capture human auditory perception. This is particularly problematic given the one-to-many nature of video-to-audio mapping, where multiple plausible soundtracks could accompany the same visuals. HarmoniDPO addresses this fundamental limitation by pioneering preference-based optimization in V2A generation, using online Direct Preference Optimization (online-DPO) to directly learn from human comparative judgments rather than reconstruction targets, thus generating audio that better aligns with perceptual quality preferences. }

\subsection{Audio-visual Datasets}
The advancement of audio generation research has been significantly propelled by the development of robust and diverse audio-visual datasets. These high-quality datasets serve as the cornerstone for both training and evaluating models, facilitating the exploration of intricate audio synthesis tasks, including text-to-audio and video-to-audio generation. Among the pioneering datasets in this domain, VGGSound \cite{chen2020vggsound} has played a pivotal role. It comprises over 200,000 video clips spanning 310 diverse audio categories, providing a rich resource for general audio-visual learning tasks. However, despite the availability of such datasets, high-quality audio-visual data remains relatively scarce, particularly for non-speech domains. This scarcity stems from the fact that the majority of existing audio-visual datasets are predominantly composed of speech-related content, which limits their applicability to broader audio generation tasks.

FineVideo represents a significant effort to address this gap, offering a high-quality open video dataset hosted on HuggingFace. This collection encompasses over 43,000 videos, totaling approximately 3,400 hours of content with an average duration of 4.7 minutes per video. Each entry includes paired audio and text annotations, making it particularly valuable for multimodal research in video-to-audio generation and audio-text alignment. Beyond FineVideo, several other datasets have made substantial contributions to audio-visual research. Ego4D \cite{grauman2022ego4d} provides a comprehensive human activity dataset containing 3,670 hours of footage, including 2,535 hours of audio content. The MultiTalk \cite{sung2024multitalk} dataset encompasses over 420,000 hours of multilingual talking head videos, while AudioSetCaps \cite{bai2024audiosetcaps} combines AudioSet, YouTube-8M, and VGGSound to offer approximately 16,000 hours of content with synthetic captions and question-answering capabilities. Additionally, AVCaps \cite{sudarsanamavcaps} contributes 28.8 hours of content with modality-specific captions, and MUSIC-AVQA \cite{li2022learning} provides approximately 150 hours of specialized music-related audio-visual content for question-answering tasks.

\subsection{Reinforcement Learning from Human Feedback}
\label{sec:bg:RLHF}
Reinforcement Learning from Human Feedback (RLHF) has emerged as a transformative paradigm for aligning AI systems with human preferences and values. Initially introduced by Christiano et al.~\cite{christiano2017deep}, RLHF was first applied to reinforcement learning tasks such as robotics and game-playing agents. The methodology was later adapted to natural language processing, where it became a cornerstone for aligning large language models (LLMs) with human intent.

Stiennon et al.\cite{stiennon2020learning} pioneered the application of RLHF in the LLM training pipeline, establishing a three-stage framework: pre-training, supervised fine-tuning (SFT), and RLHF-based alignment. This framework was further refined and scaled in subsequent work, notably InstructGPT\cite{ouyang2022training}, which demonstrated significant improvements in aligning model outputs with human preferences through RLHF-based optimization. Direct Preference Optimization (DPO)\cite{rafailov2023direct} reformulated preference learning as a binary classification problem, eliminating the need for explicit reward modeling and reducing computational complexity. Concurrently, alternative approaches such as Reward Ranked Fine-Tuning (RAFT)\cite{dong2023raft} and Rank Responses to align Human Feedback (RRHF)\cite{yuan2023rrhf} have introduced more efficient training paradigms for incorporating preference signals. Additionally, Constitutional AI\cite{bai2022constitutional} expanded the RLHF framework by incorporating explicit ethical considerations into the feedback loop, addressing safety and ethical alignment concerns. 

The success of RLHF in language models has recently inspired its application to multimodal domains. These applications can be broadly categorized into two approaches: reward-based methods and DPO-based methods. Reward-based methods like ImageReward~\cite{xu2023imagereward} train specialized reward models that capture human aesthetic and quality preferences to guide the generation process. In contrast, DPO-based methods such as Diffusion-DPO~\cite{wallace2024diffusion} adapt the DPO methodology to multimodal tasks, primarily in image generation.

However, a fundamental limitation of conventional DPO approaches is their offline nature. While standard DPO can achieve RLHF-like effects, it remains an offline reinforcement learning method where the implicit ``reward model" remains static throughout training. This limitation creates a critical distribution shift problem: as the policy model improves, it increasingly diverges from the preference model's distribution, ultimately leading to performance bottlenecks. Our work addresses this limitation by introducing the first online DPO methodology for video-to-audio (V2A) generation. Our approach achieves the adaptive benefits of online reinforcement learning while preserving the computational efficiency of DPO. This innovation is particularly valuable in the V2A domain, where no prior alignment techniques have been applied, and where the complex, context-dependent nature of audio generation demands more dynamic preference modeling.

\section{Method}
\label{sec:method}
In this section, we begin by introducing the foundational concepts of LDMs and their applications in audio generation. We then present our proposed method, \textit{HarmoniDPO (HDPO)}, which integrates video, image to enable cross-modal audio generation with enhanced performance.
\subsection{Preliminaries}
\label{sec:prelim}
\textcolor{blue}{Latent diffusion models (LDMs) \cite{rombach2022high} are a class of generative models built upon the principles of Denoising Diffusion Probabilistic Models (DDPMs) \cite{ho2020denoising}. LDMs have emerged as powerful frameworks in text-to-audio and video-to-audio generation tasks. A key feature of LDMs is that they operate in a lower-dimensional latent space, making them computationally efficient. In the context of audio, raw signals are first transformed into mel-spectrograms. The mel-spectrogram, a time-frequency representation, logarithmically scales its frequency axis to the Mel scale, which mirrors human auditory perception, enabling the model to effectively learn both temporal dynamics and spectral characteristics.}

In a video-to-audio system, given an audio signal \( x_a \) and its corresponding video \( x_v \), we first use a pre-trained variational autoencoder (VAE) with an encoder \( \mathcal{E}_{\theta} \) to compress the mel-spectrogram representation of \( x_a \) into a low-dimensional latent representation, denoted as \( z_0 = \mathcal{E}_{\theta}(x_a) \). This latent vector \( z_0 \) is the target for the diffusion model.

\textcolor{blue}{The \textbf{forward diffusion process} gradually adds Gaussian noise to the latent vector \( z_0 \) over a sequence of \( T \) timesteps. This process is a fixed Markov chain defined as:
\begin{equation}
    q(z_t|z_{t-1}) = \mathcal{N}(z_t; \sqrt{1-\beta_t}z_{t-1}, \beta_t\mathbf{I})
\end{equation}
where \( \{\beta_t\}_{t=1}^T \) is a pre-defined variance schedule. A useful property of this process is that we can sample \( z_t \) directly from \( z_0 \) at any timestep \( t \):
\begin{equation}
    q(z_t|z_0) = \mathcal{N}(z_t; \sqrt{\bar{\alpha}_t}z_0, (1-\bar{\alpha}_t)\mathbf{I})
    \label{eq:forward_process_closed_form}
\end{equation}
where $\bar{\alpha}_t = \prod_{s=1}^t(1-\beta_s)$. Using the reparameterization trick, any \(z_t\) can be expressed as \( z_t = \sqrt{\bar{\alpha}_t}z_0 + \sqrt{1-\bar{\alpha}_t}\epsilon \), where \( \epsilon \sim \mathcal{N}(0, \mathbf{I}) \).}

\textcolor{blue}{The \textbf{reverse process} aims to learn the true posterior \( q(z_{t-1}|z_t) \), which would reverse the noising process. As this posterior is intractable, a neural network \( \epsilon_{\theta} \) is trained to approximate it. As shown by Ho et al. \cite{ho2020denoising}, instead of modeling the reverse distribution's mean directly, the training can be simplified to predicting the noise component \( \epsilon \) that was added to create \( z_t \). This leads to a simplified training objective, which is a mean-squared error loss between the true and predicted noise:
\begin{equation}
    \mathcal{L}_{\text{LDM}} = \mathbb{E}_{z_0, \epsilon, t}\left[\|\epsilon - \epsilon_{\theta}(z_t, t)\|_2^2\right]
\end{equation}
To enable \textbf{conditional generation}, the denoising network \( \epsilon_{\theta} \) is augmented with conditioning information, such as a visual embedding \( f_v \), typically via a cross-attention mechanism. The objective thus becomes:
\begin{equation}
    \mathcal{L}_{\text{LDM}} = \mathbb{E}_{z_0, \epsilon, t, f_v}\left[\|\epsilon - \epsilon_{\theta}(z_t, t, f_v)\|_2^2\right]
\end{equation}}

\begin{figure*}[t]
    \centering
    \includegraphics[width=0.9\textwidth]{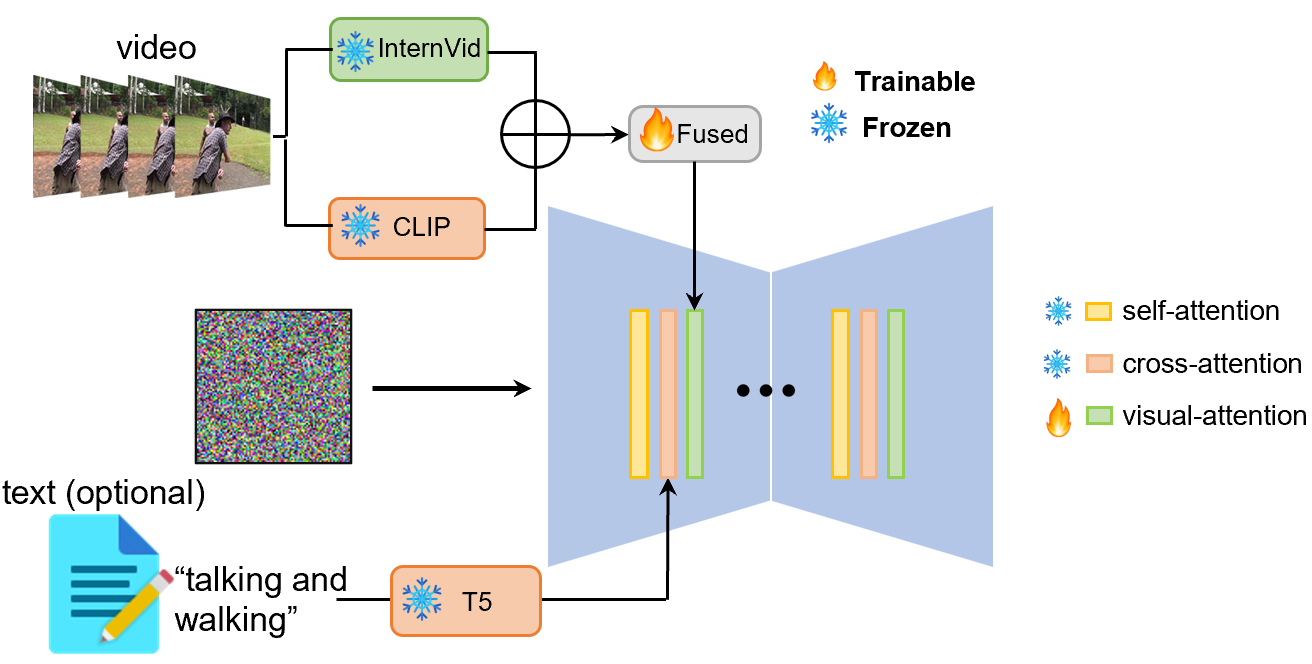}
    \caption{Illustration of our HarmoniDPO model architecture. We propose a dual-feature representation framework that combines global video features with fine-grained frame-wise features.}
    \label{fig:2}
\end{figure*}

During inference, given a visual embedding $f_v$, we can generate the corresponding audio by iterative denoising from random Gaussian noise $z_T$:
\begin{equation}
z_{t-1} = \frac{1}{\sqrt{1-\beta_t}}(z_t - \frac{\beta_t}{\sqrt{1-\bar{\alpha}t}}\epsilon_{\theta}(z_t,t,f_v))
\end{equation}
Given the limited size and varying quality of existing visual-audio datasets, we use a pre-trained text-to-audio diffusion model, Tango-2~\cite{majumder2024tango}, as our base model, keeping it frozen throughout training. Figure~\ref{fig:2} provides an overview of the framework, and we will describe each component of HarmoniDPO in detail in the following sections. 

\subsection{Visual Condition}
Accurate video representation is essential for effectively capturing the semantic content of videos. Previous methods, such as FoleyCrafter, have relied on the CLIP image encoder to obtain average semantic features from individual frames. However, since CLIP is not trained on video datasets, this approach may fall short of capturing the temporal dynamics and motion patterns inherent in video content. To address this limitation, we propose a dual-feature representation framework that combines global video features with fine-grained frame-wise features, enabling comprehensive encoding of both spatial and temporal information. 

As shown in Figure ~\ref{fig:2}, for global video feature extraction, we leverage the video encoder from InternVid~\cite{wang2023internvid}, a state-of-the-art contrastive video-language pre-training framework. Given an input video $V$ containing $T$ frames, the global video feature $f_g$ is computed as:
\begin{equation}
f_g = \mathcal{E}_{\text{InternVid}}(V) \in \mathbb{R}^{d_g}
\end{equation}
where $\mathcal{E}_{\text{InternVid}}$ represents the InternVid video encoder and $d_g$ denotes the dimensionality of the extracted video feature.
\textcolor{blue}{InternVid's architecture extends CLIP's visual understanding capabilities through continued pre-training on large-scale video-text datasets, enabling superior modeling of temporal relationships and motion dynamics across video sequences. The encoding process operates as follows: given a target video, we first establish a fixed frame sampling count $M$. The video is then temporally subsampled at $M$ uniformly distributed timestamps to ensure consistent temporal coverage. The resulting frames are concatenated and processed through a pre-trained Vision Transformer to produce the final global video representation.}

While the global video features effectively capture high-level semantic content and temporal dynamics, fine-grained frame-level details remain crucial for maintaining precise temporal alignment between the generated audio and specific visual events. Therefore, we implement a complementary frame-level processing stream that operates parallel to global feature extraction. We uniformly sample $N$ frames $\{v_1, v_2, \dots, v_N\}$ from the input video and process each frame through the CLIP visual encoder to obtain frame-wise embeddings:
\begin{equation}
e_i = \mathcal{E}_{\text{CLIP}}(v_i) \in \mathbb{R}^{d_f}, \quad i \in {1, 2, \dots, N}
\end{equation}
where $\mathcal{E}_{\text{CLIP}}$ represents the CLIP image encoder and $d_f$ is the dimension of frame features. \textcolor{blue}{In our implementation, we sample $N=250$ frames from the original video as uniformly as possible. For videos containing fewer than 250 frames, we pad the sequence with zero-filled frame features. Note that the majority of samples in our training dataset exceed 250 frames, making padding necessary for only a small portion of the data.}

For effective multimodal fusion, we first use projection layers to align the dimensions of global and frame-level features to a common dimension $d_c$:
\begin{align}
f'_g &= W_g f_g \in \mathbb{R}^{d_c} \\
e'_i &= W_f e_i \in \mathbb{R}^{d_c}
\end{align}
where $W_g \in \mathbb{R}^{d_c \times d_g}$ and $W_f \in \mathbb{R}^{d_c \times d_f}$ are learnable projection matrices. We then concatenate the projected global video feature with the projected frame-wise embeddings:
\begin{equation}
F = [f'_g; e'_1; e'_2; \dots; e'_N] \in \mathbb{R}^{(1+N) \times d_c}
\end{equation} Finally, we employ another projection layer to transform the concatenated features to the model's hidden dimension:
\begin{equation}
F' = W_h F \in \mathbb{R}^{(1+N) \times d_h}
\end{equation}
where $W_h \in \mathbb{R}^{d_h \times d_c}$ is the final projection matrix and $d_h$ is the hidden dimension of the model.

\textcolor{blue}{To better capture temporal relationships among the fused features, we incorporate Rotary Position Embeddings (RoPE)~\cite{su2024roformer} within a multi-head self-attention mechanism. The self-attention operation employs separate learned projections for queries, keys, and values:
\begin{align}
Q = F' W_q \in \mathbb{R}^{(1+N) \times d_h}, \
K = F' W_k \in \mathbb{R}^{(1+N) \times d_h}, \
V = F' W_v \in \mathbb{R}^{(1+N) \times d_h}
\end{align}
where $W_q, W_k, W_v \in \mathbb{R}^{d_h \times d_h}$ are learnable projection matrices for queries, keys, and values, respectively. The final fused representation is obtained by:
\begin{equation}
F_{\text{fused}} = \text{Attention}_{\text{RoPE}}(Q, K, V) \in \mathbb{R}^{(1+N) \times d_h}
\end{equation}
where RoPE is applied to both queries $Q$ and keys $K$ before computing attention scores, enabling the model to encode sequential ordering of frames through relative position information. These fused representations are subsequently incorporated into the diffusion model via a cross-attention layer (visual-attention), enabling the denoising process to dynamically attend to spatio-temporal features during training. }

\begin{figure*}[t]
    \centering
    \includegraphics[width=0.95\textwidth]{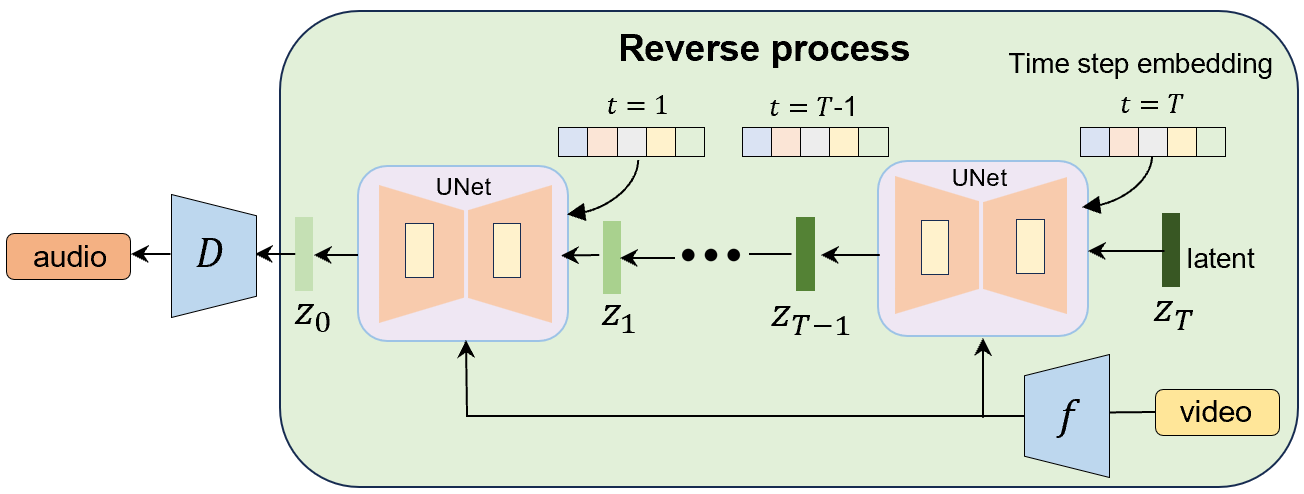}
    \caption{\textcolor{blue}{Overview of the video-conditioned audio generation pipeline. The diagram illustrates the reverse diffusion process, where noisy latent variables $z_t$ are iteratively denoised through a U-Net architecture. }}
    \label{fig:2.2}
\end{figure*}

\textcolor{blue}{Building upon the comprehensive visual, our method integrates these multimodal conditions into a diffusion-based audio generation framework. As illustrated in Figure ~\ref{fig:2.2}, the generation process operates within the latent space of a pre-trained Variational Autoencoder (VAE) for computational efficiency. The reverse diffusion process begins with noisy latent variables $z_T$ and iteratively denoises them through a U-Net architecture, where the fused visual representations $F_{\text{fused}}$ are injected via cross-attention layers to guide the denoising process. The final denoised latent $z_0$ is then decoded through the VAE decoder to generate the target audio output.}

Our model also supports the optional textual conditioning for enhanced control over the generation process. For the \textbf{Text} input, we use the pre-trained FLAN-T5\cite{chung2024scaling} text encoder to extract the text embedding $f_t$. Given a text prompt $T$, the text embedding $f_t$ is computed as:  
\begin{equation}  
f_t = \mathcal{E}_{\text{T5}}^{\text{text}}(T) \in \mathbb{R}^{d_t}  
\end{equation}  
where $\mathcal{E}_{\text{T5}}^{\text{text}}$ denotes the FLAN-T5 text encoder and $d_t$ is the dimension of the text feature. We use a projection layer to match the hidden dimension. The diffusion model then integrates these textual conditions through the cross-attention layer. 

\subsection{Direct Preference Optimization }
Recent advances in preference-based optimization have demonstrated the effectiveness of Reinforcement Learning from Human Feedback (RLHF) for aligning large language models (LLMs). However, its application to cross-modal generation tasks, such as video-to-audio synthesis, remains underexplored. In this paper, We propose the first online-DPO framework for cross-modal preference alignment in video-conditioned audio generation. Unlike offline DPO that requires pre-collected static datasets, our method continuously adapts to human preferences through real-time interaction during the generation process. 

\subsubsection{Theoretical Foundation}
\label{ssec:dpo_foundation}

Direct Preference Optimization (DPO) establishes a direct mapping between reward functions and optimal policies through three fundamental principles:

\begin{enumerate}
    \item \textbf{Optimal Policy Characterization}: \textcolor{blue}{For any reward function $r(v,a)$ and reference policy $\pi_{\text{ref}}$, the KL-constrained optimal policy has closed-form solution:
    \begin{equation}
        \pi^*(a|v) = \frac{\pi_{\text{ref}}(a|v)\exp(r(v,a)/\beta)}{Z(v)}
        \label{eq:optimal_policy}
    \end{equation}
    where $Z(v) = \sum_a \pi_{\text{ref}}(a|v)\exp(r(v,a)/\beta)$ is the partition function, $v$ and $a$ are the inputs and generated outputs. In our paper, $v$ represents the input video and $a$ represents the output audio.  }

    \item \textbf{Reward-Policy Duality}: Equation~\ref{eq:optimal_policy} permits reward reparameterization via policy ratios:
    \begin{equation}
        r(v,a) = \beta\log\frac{\pi^*(a|v)}{\pi_{\text{ref}}(a|v)} + \beta\log Z(v)
        \label{eq:reward_dual}
    \end{equation}

    \item \textbf{Preference Modeling}: \textcolor{blue}{Human judgments follow the Bradley-Terry model:
    \begin{equation}
        p^*(a_w \succ a_l|v) = \frac{\exp(r(v,a_w))}{\exp(r(v,a_w)) + \exp(r(v,a_l))}
        \label{eq:preference_model}
    \end{equation}
    where $a_w$ denotes the preferred output over $a_l$ for a given input $v$. }
\end{enumerate} 

Substituting~\ref{eq:reward_dual} into~\ref{eq:preference_model} yields the DPO objective:
\begin{equation}
    \mathcal{L}_{\text{DPO}} = -\mathbb{E}_{(v,a_w,a_l)\sim\mathcal{D}}\left[
    \log \sigma\left(
        \beta\log\frac{\pi_\theta(a_w|v)}{\pi_{\text{ref}}(a_w|v)} - 
        \beta\log\frac{\pi_\theta(a_l|v)}{\pi_{\text{ref}}(a_l|v)}
    \right)
    \right]
    \label{eq:dpo_loss}
\end{equation}

where $\sigma(\cdot)$ denotes the sigmoid function, and $\mathcal{D}$ is the preference dataset.

\subsubsection{Online-DPO}
While DPO provides an elegant framework for preference optimization, its conventional offline formulation faces fundamental scalability challenges. The approach critically depends on large-scale human-annotated preference data, which becomes prohibitively expensive to collect for complex cross-modal tasks like video-to-audio generation—where expert annotators must evaluate subtle audiovisual synchronization and acoustic quality. This dependency creates a bottleneck for adapting to new domains or evaluation criteria, as each requires fresh annotation campaigns. Moreover, the static nature of offline datasets leads to inevitable obsolescence: as generative models improve, human judgments collected on older model outputs become misaligned with current capabilities, causing the preference model to optimize for outdated behaviors. This problem is exacerbated in dynamic multi-modal settings where aesthetic standards and technical expectations evolve rapidly, rendering pre-collected preferences increasingly irrelevant over time. 

\begin{figure*}[t]
    \centering
    \includegraphics[width=0.95\textwidth]{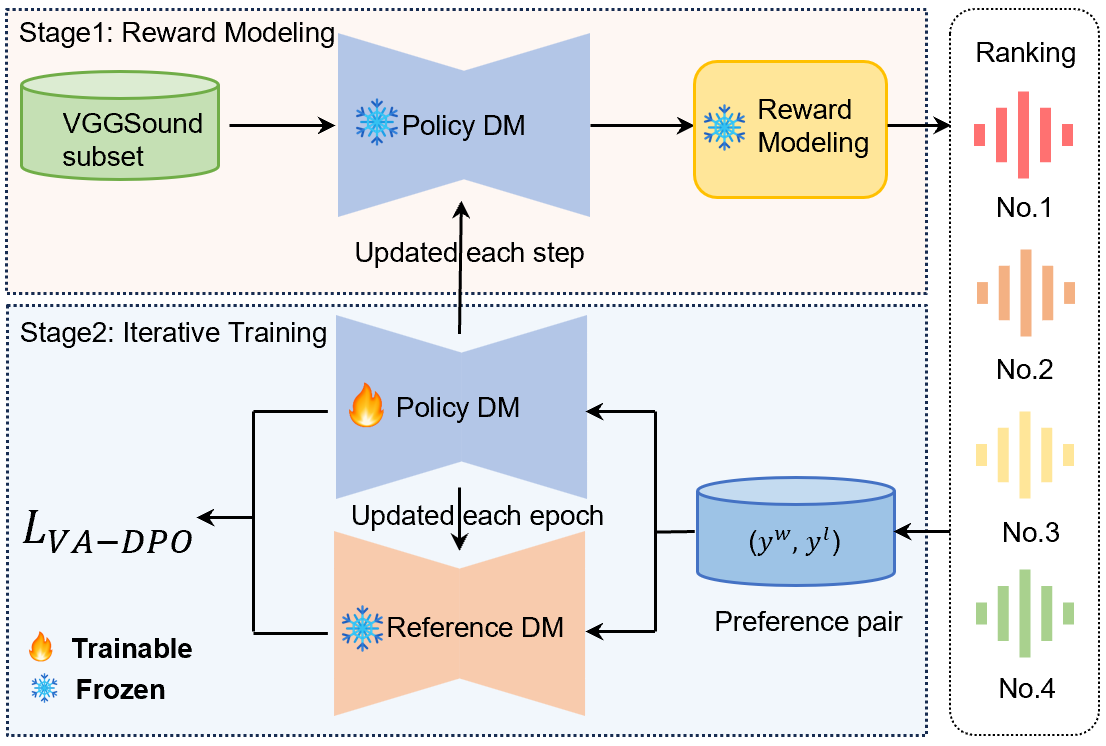}
    \caption{Pipeline of one Online-DPO process. We use a subset of VGGSound to fine-tune the whole model.}
    \label{fig:3}
\end{figure*}

Inspired by recent work \cite{wang2024self}, we propose an Online-DPO method which can explore an iterative self-training approach and uses no human-annotated preferences in the training loop. As illustrate in Figure \ref{fig:3}, our onlin-DPO mainly consists of two parts: 1) areward modeling, and 2) iterative training using the modeled rewards.

\textbf{Reward Modeling}
Building upon our motivation for preference-based optimization, we develop an automated reward modeling framework that addresses the key limitations of reconstruction-based training identified earlier. Recognizing that human preference annotation at scale is prohibitively expensive while raw audio-visual pairs may not reflect true perceptual quality, we propose a multi-dimensional reward system that captures the essential aspects of human evaluation. Our approach evaluates generated audio through three critical lenses that mirror human judgment criteria: (1) temporal and semantic alignment with visual content, (2) consistency with textual descriptions (when available), and (3) intrinsic acoustic quality. This tripartite evaluation directly addresses the ambiguity and subjectivity challenges highlighted in our introduction, where multiple audio outputs could technically match the video but vary significantly in perceptual quality. 

Specifically, we assess the quality of the generated audio along three primary dimensions:
\begin{itemize}
    \item \textbf{Audio-Visual Correspondence:} The temporal and contextual alignment between the synthesized audio and the input video content.
    \item \textbf{Audio-Text Consistency:} The accuracy with which the audio reflects accompanying textual descriptions.
    \item \textbf{Intrinsic Audio Quality:} The acoustic fidelity and clarity of the generated audio itself, irrespective of the conditioning inputs.
\end{itemize}

To operationalize this multi-faceted evaluation, we utilize specific automated metrics for each dimension. Audio-visual correspondence is quantified via the similarity score obtained from the pre-trained contrastive learning model CAV-MAE~\cite{gong2022contrastive}. Similarly, audio-text consistency is assessed using the similarity score derived from a dedicated audio-text contrastive learning model CLAP. Most crucially, we employ the recently proposed Audiobox-Aesthetics predictor from Meta~\cite{tjandra2025meta}. This predictor provides non-intrusive, utterance-level aesthetic scores based on human perceptual dimensions, trained across diverse audio types (speech, music, sounds), offering a robust measure of perceptual quality suitable for generative models.

Given that the three reward components—audio-visual correspondence similarity ($R_{av}$), audio-text consistency similarity ($R_{at}$), and the intrinsic audio quality score ($R_{quality}$) from Audiobox-Aesthetics—operate on potentially different scales and represent distinct facets of audio generation quality, a direct summation is inadequate for creating a unified reward signal. To address this, we first normalize each component independently to ensure they contribute appropriately to the final reward. The final composite reward score $R(y)$ for a generated audio sample $y$ is then calculated as a weighted linear combination of these normalized scores:
\begin{equation}
    R(y) = w_{av} \cdot R'_{av}(y) + w_{at} \cdot R'_{at}(y) + w_{quality} \cdot R'_{quality}(y) 
\end{equation}
Here, $w_{av}$, $w_{at}$, and $w_{quality}$ are non-negative hyperparameters representing the relative importance assigned to audio-visual correspondence, audio-text consistency, and intrinsic quality, respectively. 


\textbf{Iterative Training with Value Aware DPO Loss}

A key limitation of conventional offline DPO is its reliance on a static preference dataset. This dataset, collected prior to training, prevents the policy from receiving direct feedback on the quality variations within its own generated outputs during the alignment process. Consequently, the model may not efficiently learn to refine subtle aspects captured by the reward function, leading to suboptimal performance. To overcome this challenge, our Online-DPO framework implements an Online Preference Sample Generation strategy integrated within the iterative training loop.

Specifically, instead of drawing from a fixed dataset, preference pairs are generated dynamically in each training iteration $t$. For a given input condition $x$ (video and text), the current policy $\pi_t$ generates a set of $N$ candidate audio samples, denoted as $\mathcal{Y} = \{y_1, y_2, ..., y_N\}$. Each candidate $y_i \in \mathcal{Y}$ is subsequently evaluated using our composite automated reward function $R(y_i)$ 

Crucially, we then identify the best-performing ($y_w$) and worst-performing ($y_l$) samples within this generated set $\mathcal{Y}$ based purely on the automated reward scores:
\begin{equation} \label{eq:best_worst_selection}
y_w = \underset{y_i \in \mathcal{Y}}{\operatorname{argmax}} R(y_i) 
\quad \text{and} \quad 
y_l = \underset{y_i \in \mathcal{Y}}{\operatorname{argmin}} R(y_i) 
\end{equation}
These dynamically generated preference pairs $\{x, y_w, y_l\}$ form the online dataset $\mathcal{D}_t$ used to update the policy $\pi_\theta$.

To further enhance learning efficiency and sensitivity to the magnitude of preference, we propose a \textbf{Value-Aware DPO (VA-DPO)} loss that explicitly incorporates reward magnitudes. Unlike standard DPO, which treats all preference pairs equally (as a binary choice), our loss accounts for the \textit{degree} of preference between samples by integrating the reward margin. The resulting VA-DPO loss is defined as:
\begin{equation} \label{eq:va_dpo_loss}
\begin{aligned}
\mathcal{L}_{\text{VA-DPO}}(\pi_\theta; \pi_{\text{ref}}) =
-\mathbb{E}_{\substack{(x,y_w,y_l)\\ \sim\mathcal{D}_t}} \bigg[ & \log \sigma \bigg( \beta \log \tfrac{\pi_\theta(y_w|x)}{\pi_{\text{ref}}(y_w|x)}
- \beta \log \tfrac{\pi_\theta(y_l|x)}{\pi_{\text{ref}}(y_l|x)} \\
& \qquad - \lambda \big(R(y_w) - R(y_l)\big) \bigg) \bigg]
\end{aligned}
\end{equation}
where $\sigma(\cdot)$ is the sigmoid function, $\beta$ is the temperature parameter controlling the weight of the policy difference, and $\lambda$ is a hyperparameter balancing the DPO term and the reward margin term. \textcolor{blue}{Considering that diffusion models are not like Large Language Models(LLMs), we follow Diffusion-DPO\cite{majumder2024tango,wallace2024diffusion} for the loss calculation. Regarding the reference model $\pi_{\text{ref}}$, we update it periodically; specifically, at the end of each training epoch in our implementation.
}

\subsection{Test-time Scaling}
\textcolor{blue}{To enhance the inference performance of our video-to-audio diffusion model, we propose a Dual-scale Diffusion Search (DDS) strategy. This approach introduces a dual-scale sampling mechanism that adaptively explores the latent space by combining both small and large step sizes, thereby balancing local exploitation and global exploration.}

The DDS algorithm maintains a population of candidate solutions that evolve over multiple iterations.  Initially, the population $P_0$ is sampled from a standard normal distribution.  At each iteration, for every candidate (generated audios) in the population, we generate two potential updates using different diffusion scales: a conservative step with parameter $\beta_s$ for local exploitation and a more aggressive step with parameter $\beta_l$ for global exploration $(\beta_l > \beta_s)$. Specifically, given a candidate solution, we generate two new candidates:
\begin{equation}
    x_s = \beta_s x + \sqrt{1 - \beta_s^2} \cdot \eta
\end{equation}
\begin{equation}
    x_l = \beta_l x + \sqrt{1 - \beta_l^2} \cdot \eta
\end{equation}
where $\eta$ is sampled from $\mathcal{N}(0, 1)$. 

To evaluate the quality of the generated audio, we employ a zero-shot evaluation approach.  Specifically, we utilize the CLIP score \cite{wu2022wav2clip} to measure the audio-visual correspondence and the CLAP score to assess the audio-text alignment.  These metrics provide a comprehensive evaluation of the generated audio in terms of both visual and textual relevance. The complete DSS procedure is presented in Algorithm~\ref{alg1}

\begin{algorithm}
\caption{Dual-scale Diffusion Search}
\begin{algorithmic}
\Require $D, T, N, \beta_s, \beta_l, \mathcal{F}$
\Ensure best\_solution, best\_score
\State Initialize $P_0$ with $N$ samples from $\mathcal{N}(0, 1)$
\State $\text{best\_score} \Leftarrow -\infty$
\State $\text{best\_solution} \Leftarrow \emptyset$
\For{$i \Leftarrow 1$ to $N$}
    \If{$\mathcal{F}(P_0[i]) > \text{best\_score}$}
        \State $\text{best\_score} \Leftarrow \mathcal{F}(P_0[i])$
        \State $\text{best\_candidate} \Leftarrow P_0[i]$
    \EndIf
\EndFor
\For{$t \Leftarrow 1$ to $T$}
    \For{$i \Leftarrow 1$ to $N$}
        \State Sample $\eta$ from $\mathcal{N}(0, 1)^D$
        \State $x_s \Leftarrow \beta_s P_{t-1}[i] + \sqrt{1 - \beta_s^2} \cdot \eta$
        \State $x_l \Leftarrow \beta_l P_{t-1}[i] + \sqrt{1 - \beta_l^2} \cdot \eta$
        \If{$\mathcal{F}(x_s) > \mathcal{F}(x_l)$}
            \State $P_t[i] \Leftarrow x_s$
            \If{$\mathcal{F}(x_s) > \text{best\_score}$}
                \State $\text{best\_score} \Leftarrow \mathcal{F}(x_s)$
                \State $\text{best\_candidate} \Leftarrow x_s$
            \EndIf
        \Else
            \State $P_t[i] \Leftarrow x_l$
            \If{$\mathcal{F}(x_l) > \text{best\_score}$}
                \State $\text{best\_score} \Leftarrow \mathcal{F}(x_l)$
                \State $\text{best\_candidate} \Leftarrow x_l$
            \EndIf
        \EndIf
    \EndFor
\EndFor
\end{algorithmic}
\label{alg1}
\end{algorithm}

\section{Experiments}\label{sec2}
\paragraph{Dataset} To train our proposed HarmoniDPO, we utilize the VGGSound~\cite{chen2020vggsound} dataset, adhering to its original train/test splits. For evaluation, we assess our method using both the VGGSound and AVSync15~\cite{zhang2024audio} datasets.

\paragraph{Implementation Details}
\textcolor{blue}{We use Tango-2~\cite{majumder2024tango}, a pre-trained text-to-audio diffusion model as our based diffusion model. For visual feature extraction, we utilize OpenCLIP ViT-H/14~\cite{cherti2023reproducible} as our image encoder and ViCLIP-L-14 as our video encoder, where the latter is pre-trained on the InternVid-10M-FLT\cite{wang2023internvid} dataset. }

\textcolor{blue}{During training, We use 64 H800 GPUs for experiments. We employ the AdamW optimizer with a constant learning rate of \(1 \times 10^{-4}\) and a batch size of 128 visual-audio embedding pairs. To enhance model generalization, we apply classifier-free guidance with a dropout rate of 0.05. The training process consists of two distinct phases: (1) In the visual conditioning phase, we keep the pre-trained diffusion model frozen while fine-tuning only the newly added projection head and cross-attention layers to adapt the text-to-audio diffusion model into a video-to-audio model; (2) In the subsequent alignment stage, we first sample 4500 videos independently and identically distributed (i.i.d.) from the VGGSound dataset. For the online-DPO fine-tuning, we generate candidate audio samples for these videos. To ensure the generated candidates are of high quality and relevance, we condition the generation on the original text descriptions associated with each video. Furthermore, inspired by the methodology in Tango-2, we create diverse candidate sets by varying parameters such as the number of diffusion steps and the classifier-free guidance scale during generation. We set $w_{av}$, $w_{at}$ and $w_{quality}$ to be equal (1) for the reward modeling stage, adopting a balanced combination of audio-visual alignment, audio-text relevance, and perceptual quality scores. Finally, we perform end-to-end fine-tuning of the entire model using the online-DPO method, leveraging these generated candidates to learn preference alignments. This two-phase approach ensures stable training while effectively aligning the visual and audio modalities. }

\paragraph{Metrics} To evaluate the performance, we employ a comprehensive set of metrics that assess various aspects of the generated audio, including: Mean KL Divergence (MKL)\cite{iashin2021taming} to measure the sample-level similarity between the generated audio and the ground truth; CLIP Similarity to evaluate the semantic coherence between the input video and the generated audio embeddings, using Wav2CLIP\cite{wu2022wav2clip} as the audio encoder and CLIP as the video encoder, as done in previous works~\cite{wang2024v2a,zhang2024foleycrafter}; \textcolor{blue}{Frechet Inception Distance (FID)}\cite{heusel2017gans} and Frechet Audio Distance (FAD)\cite{kilgour2018fr} with VGGish\cite{hershey2017cnn} to assess the fidelity and distribution similarity of the generated audio; CLAP Similarity to evaluate the cross-modal alignment between text and generated audio; and Onset Acc (onset detection accuracy) and Onset AP (onset detection average precision) \cite{xie2024sonicvisionlm} to evaluate the generated audios, using the onset ground truth from the datasets. 

\paragraph{Baseline} 
For comparison, we use current state-of-the-art method SpecVQGAN~\cite{iashin2021taming}, Diff-Foley~\cite{luo2024diff}, V2A-Mapper~\cite{wang2024v2a}, Seeing-and-hearing~\cite{xing2024seeing} and Foley-crafter~\cite{zhang2024foleycrafter}. SpecVQGAN generates audio tokens autoregressively from video tokens, while Diff-Foley applies contrastive learning to achieve synchronized audio synthesis via its CAVP encoders.  V2A-Mapper aligns image representations with audio embeddings in CLAP space, facilitating video-based audio generation through a pre-trained model.  Seeing-and-hearing uses ImageBind~\cite{girdhar2023imagebind} as a connector between visual and audio domains, enabling multimodal generation by integrating existing audio and video generators. FoleyCrafter introduces both semantic and temporal adapters to enhance video-to-audio generation.

\subsection{Qualitative Evaluation}
\begin{table}[ht]
\caption{Comparison of results on VGGSound dataset. The results show that our method can achieve better results with less data. The number underlined indicates the second best result.}
\label{tab:1}
\centering
\setlength{\tabcolsep}{0.8pt} 
\begin{tabular*}{\textwidth}{@{\extracolsep{\fill}}lcccccc}
\toprule
Method & Training Set & MKL $\downarrow$ & CLIP $\uparrow$ & FD$\downarrow$ & FAD $\downarrow$& CLAP $\uparrow$ \\
\midrule
SpecVQGAN & VGGSound, VAS & 3.40 & 5.88 & 32.01 & 5.79 & - \\
Diff-Foley & VGGSound, AudioSet-V2A & 3.32 & 9.17 & 29.03 & 6.23 & 20.5 \\
V2A-Mapper & VGGSound, Land-scape & 2.84 & 9.72 & 24.16 & \textbf{1.34} & 24.5 \\
FoleyCrafter & VGGSound, AudioSet Strong & 2.56 & 10.70 &  19.67 &  2.78 & 25.3 \\
Fieren & VGGSound & 2.58 & 11.83 &12.48 &3.32& 24.7 \\
\midrule
HarmoniDPO(w/o alignment) & VGGSound & 2.04 & 12.54 & 7.72 & 1.75 & 28.35 \\
HarmoniDPO (aligned) & VGGSound & \underline{1.86} & \underline{13.27} & \underline{6.54} & 1.64 & \underline{31.75} \\
HarmoniDPO (aligned + DDS) & VGGSound & \textbf{1.82} & \textbf{13.65} & \textbf{6.42} & \underline{1.59} & \textbf{32.57} \\
\botrule
\end{tabular*}
\end{table}

\paragraph{Audio quality and cross-modal alignment }

We evaluate HarmoniDPO's performance in audio generation quality and cross-modal alignment using the VGGSound dataset and AVSync-15 dataset, with quantitative results presented in Table~\ref{tab:1} and Table~\ref{tab:2}. To ensure a fair comparison against baselines that only use visual input, we evaluate HarmoniDPO without leveraging text conditioning, thus avoiding any potential advantage from multimodal inputs.

For the VGGSound dataset, our evaluation specifically examines three progressively enhanced configurations of HarmoniDPO: 
1) The base model utilizing pre-trained video and fine-grained frame embeddings without explicit alignment (HarmoniDPO (w/o alignment)).
2) The model enhanced with DPO-based alignment (HarmoniDPO (aligned)).
3) The DPO-aligned model further enhanced with Dual-scale Diffusion Search (DDS) during inference, where we set the number of search candidates N to 4 in our experiments.

Our experimental results demonstrate significant performance improvements across these configurations. The base model already surpasses baseline methods by effectively leveraging pre-trained video embeddings and frame-level visual features. The DPO-aligned version shows marked improvements in audio-visual synchronization, particularly evident in CLIP and CLAP scores. The full model with DDS inference achieves optimal performance, maintaining competitive FAD scores while significantly outperforming all baselines in other metrics.

\begin{table*}[ht]
\caption{Comparison of results on AVSync15 dataset.}
  \centering
\setlength{\tabcolsep}{1pt}
\begin{tabular*}{\textwidth}{@{\extracolsep{\fill}}lccccc}
    \toprule
    Method & Training Set  &MKL $\downarrow$ & CLIP $\uparrow$ & FID $\downarrow$\\
    \midrule
    SpecVQGAN & VGGSound, VAS &3.603 & 6.474 & 75.56 \\
    Diff-Foley & VGGSound, AudioSet-V2A & 1.963 & 10.38 & 65.77 \\
    Seeing and Hearing & VGGSound, Land-scape  & 2.547 & 2.033 & 65.82 \\
    FoleyCrafter & VGGSound, AudioSet Strong & 1.479 & 11.94 & 36.80\\
    HarmoniDPO (Ours) & VGGSound  & \textbf{1.332} & \textbf{14.38}& \textbf{31.59} \\
    \bottomrule
  \end{tabular*}
  \label{tab:2}
\end{table*}

For the AVSync-15 dataset, we evaluate all methods on the test split, which consists of 150 distinct examples.  The AVSync-15 dataset, curated from VGGSound's most temporally challenging samples, serves as our benchmark for evaluating synchronization capabilities. We include a time alignment experiment in the subsequent analysis and limit our evaluation to three fundamental metrics. As shown in Table~\ref{tab:2}, our HarmoniDPO method achieves competitive or superior performance across all metrics.  Specifically, HarmoniDPO excels in CLIP match scores and FID, highlighting its ability to generate audio-visual content that is both semantically aligned and visually coherent.

\paragraph{Time alignment}
To evaluate temporal synchronization, we conduct experiments on the AVSync15 dataset.  This dataset is constructed from high-quality videos sourced from VGGSound, which have been carefully filtered and segmented to retain only the most precise video-audio pairs while discarding irrelevant or inactive segments.  As shown in Table~\ref{tab:3}, our method achieves state-of-the-art performance in temporal synchronization, demonstrating its effectiveness in aligning audio with visual events.

\begin{table}[h]
\caption{Comparison of results on AVSync15 dataset.}\label{tab1}%
\setlength{\tabcolsep}{1pt}
\centering
\begin{tabular}{@{}lll@{}}
\toprule
Method & Onset ACC $\uparrow$ & Onset AP $\uparrow$ \\
\midrule
SpecVQGAN &26.74 & 63.18  \\
    Diff-Foley &  21.18 & 66.55 \\
    Seeing and Hearing  & 20.95 &  60.33 \\
    FoleyCrafter & 28.48 & 68.14\\
    HarmoniDPO (Ours)  & \textbf{32.53} & \textbf{69.97} \\
\botrule
\label{tab:3}
\end{tabular}
\end{table}



\subsection{Online-DPO alignment}

\begin{table}[ht]
\caption{Comparison of results on VGGSound dataset. To enhance the robustness of the conclusions, we report the results without using DDS method.}
\label{tab:dpo1}
\centering
\setlength{\tabcolsep}{0.8pt} 
\begin{tabular*}{\textwidth}{@{\extracolsep{\fill}}lccccc}
\toprule
Num  & MKL $\downarrow$ & CLIP $\uparrow$ & FID$\downarrow$ & FAD $\downarrow$& CLAP $\uparrow$ \\
\midrule
2 & 1.97 & 12.33 & 8.34 & 1.77 & 29.35 \\
4 & \textbf{1.83} & 12.95 & 7.76 & 1.73 & 30.54 \\
8  & $1.856 \pm 0.04$ & $\textbf{13.27} \pm 0.132$ &  $\textbf{6.54} \pm 0.054$ &  $\textbf{1.64} \pm 0.013$ & $31.75 \pm 0.155$ \\
12  & 1.847 & 13.15 &  6.67 &  1.65 & \textbf{32.15} \\
\botrule
\end{tabular*}
\end{table}

\paragraph{Number of candidates} Having established the general effectiveness of our online-DPO approach (as detailed in Table~\ref{tab:1}). We now delve deeper into optimizing its performance by examining the influence of the number of candidate audios generated by our HarmoniDPO model. This hyperparameter dictates how many potential audio samples are evaluated during each step of the online alignment process. 

Our experiments on the VGGSound dataset (Table~\ref{tab:dpo1}) reveal that increasing the number of candidates generally enhances model performance, though the optimal setting varies across metrics. Specifically, performance improves significantly when scaling from two to eight candidates, with CLIP, FID, and FAD metrics achieving their best results at this point. Interestingly, we observe metric-specific variations: MKL peaks with four candidates, while CLAP scores continue to improve up to twelve candidates. 

\textcolor{blue}{This divergence across metrics suggests they measure different aspects of audio quality. For example, the continued improvement in CLAP with 12 candidates—while FID and FAD slightly worsen—indicates that the model can generate audio that better matches the semantic content but may have subtle acoustic imperfections. In other words, higher CLAP scores do not always mean better overall audio: some outputs may ``sound right" in context but contain unnatural textures or background noise. Conversely, low FID/FAD indicates realistic sound statistics but doesn’t guarantee perfect semantic alignment. Given this trade-off, we find the 8-candidate setting achieves a good balance, performing well on both types of metrics. }

\paragraph{User study results}
\textcolor{blue}{We conducted a comprehensive human evaluation of our Online-DPO framework using the AVSync15 test set, comparing four configurations: baseline (no alignment), and alignment with 4, 8, and 12 candidate audios. To ensure evaluation robustness while mitigating subjective bias, we recruited eight evaluators from our laboratory to participate in two rounds of assessments. In each round, evaluators scored two randomly selected videos per category using replacement sampling. To ensure adequate video coverage while maintaining methodological rigor, we implemented a controlled repetition mechanism: each evaluator was guaranteed to rate at least three unique videos per category, with precisely one repeated video across the two rounds. This approach yielded four total ratings per evaluator-category pair (three unique videos plus one repeated video), achieving 75\% unique sample coverage while permitting controlled reliability assessment through the intentional repetition. }

We evaluate the generated audio outputs along two critical dimensions: i) Overall Quality (OVL)\cite{liu2023audioldm} and ii) Relevance to the Input Video (REL)\cite{liu2023audioldm}. Evaluators provided Likert-scale ratings (1-5) for each sample, with higher scores indicating better performance. We tested 

\begin{figure}[t]
\centering
\includegraphics[width=0.95\linewidth]{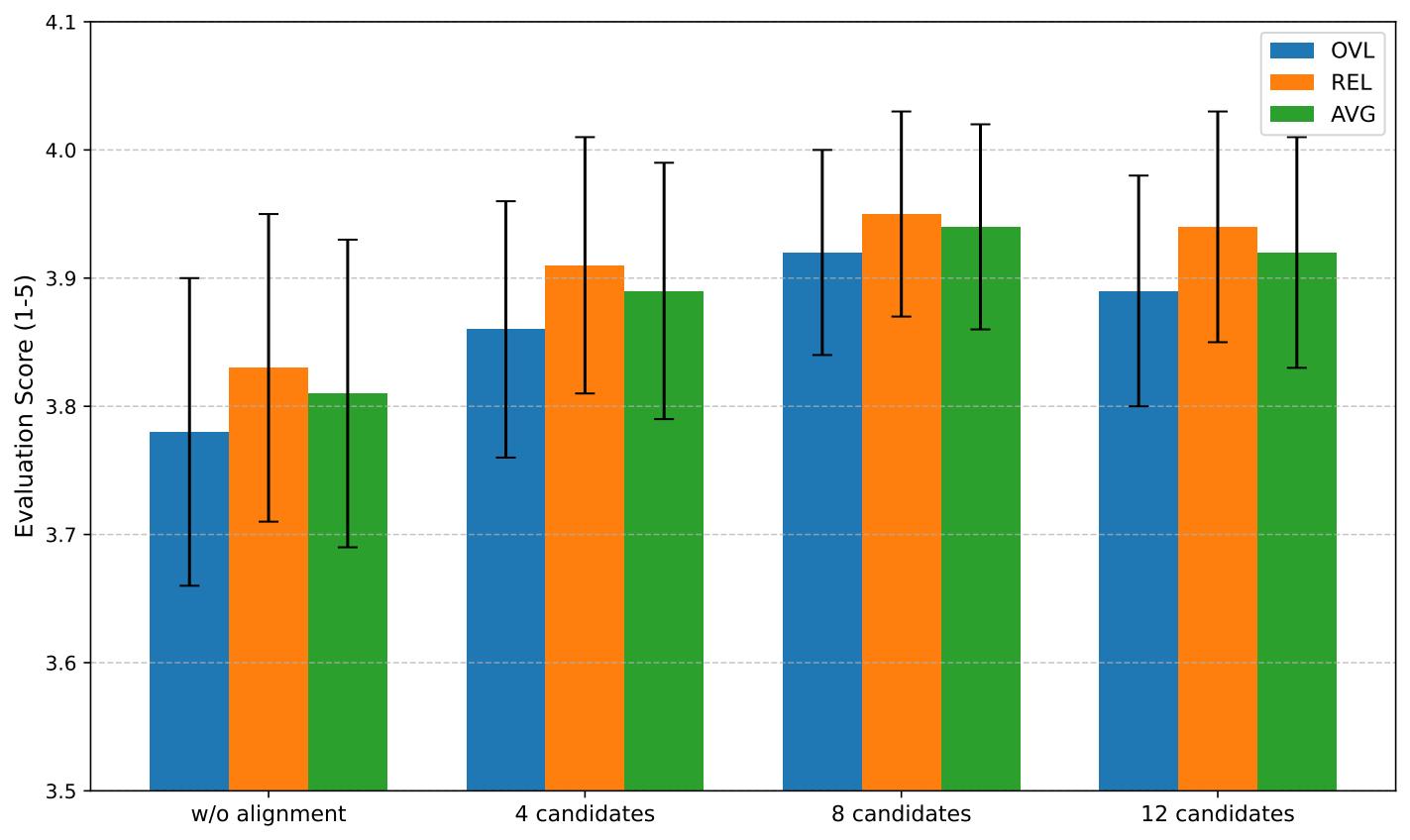}
\caption{\textcolor{blue}{Comparative analysis of user evaluation scores across HarmoniDPO configurations. The grouped bars represent Overall Quality (OVL), Video-Audio Relevance (REL), and their Average (AVG) scores, with error bars indicating evaluation consistency. Results show optimal performance at 8 candidates. Candidates are the generated audio samples.}}
\label{fig:user_study}
\end{figure}

As illustrated in Figure~\ref{fig:user_study}, we observe consistent improvements in both audio quality and video-audio synchronization across different candidate configurations. The results demonstrate that HarmoniDPO with 8 candidates achieves the best balance, particularly excelling in video-audio relevance (REL) with a score of 3.95, while maintaining strong overall quality (OVL) at 3.92.

\paragraph{Visualization Results}
\begin{figure}[t]
\centering
\includegraphics[width=0.95\linewidth]{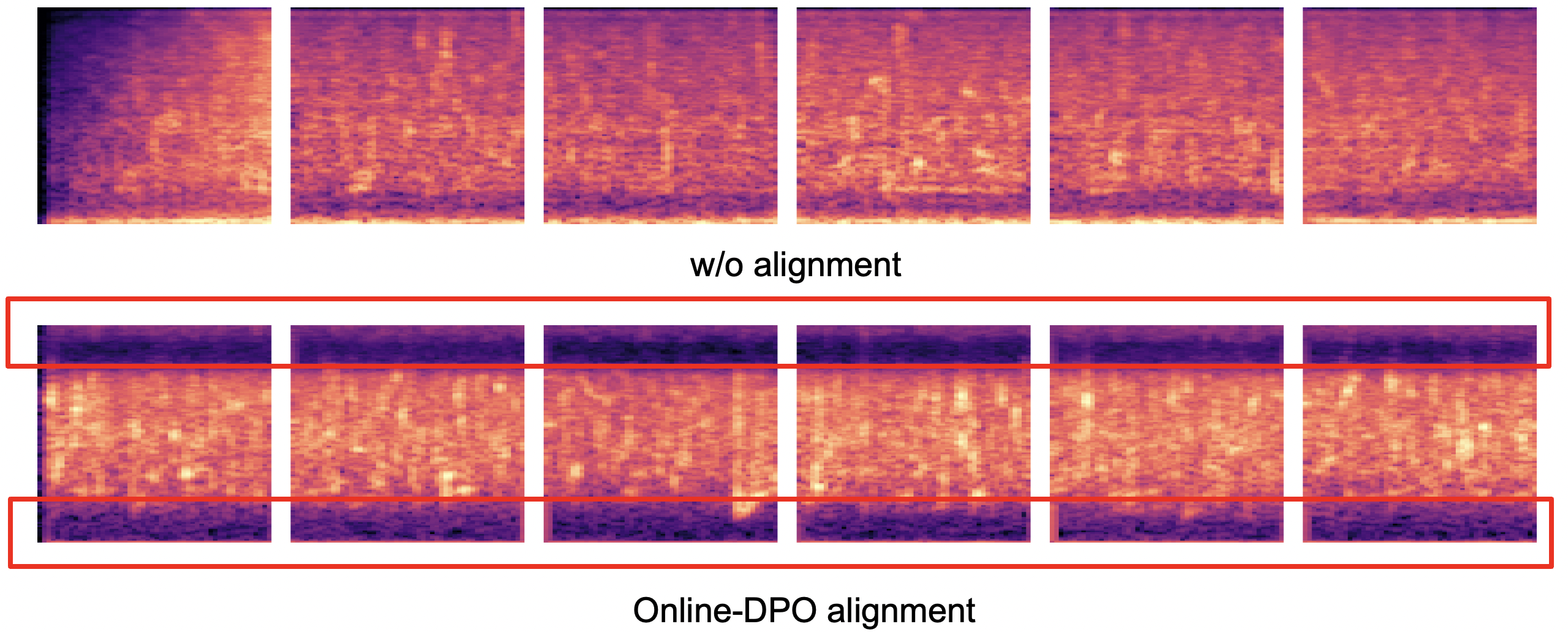}
\caption{\textcolor{blue}{Comparative spectrogram analysis between our base model (without Online-DPO alignment) and the Online-DPO-enhanced version, evaluated on a 10-second unseen valley stream video. Using Online-DPO method can effectively attenuate ambient noise while optimizing frequency response for mid-range enhancement, resulting in superior noise reduction and vocal clarity.}}
\label{fig:vis1}
\end{figure}

To validate the benefits of our proposed Online-DPO alignment module, we conduct an ablation study comparing two variants of our model: (1) the original implementation without Online-DPO alignment (serving as baseline), and (2) the enhanced version incorporating Online-DPO. As illustrated in Figure~\ref{fig:vis1}, we evaluate both versions on a 10-second valley stream video excluded from training data (non-VGGSound source).

The baseline model (without Online-DPO) generates audio with three notable characteristics: first, it exhibits uniformly strong amplitudes across all frequency bands; second, this results in perceptually louder output with less controlled spectral distribution; third, while maintaining basic audiovisual correspondence, the audio quality appears noisier and less refined.

In contrast, the Online-DPO-enhanced version demonstrates significant improvements: (a) the spectral energy becomes more focused in the mid-frequency range, (b) both high and low frequency extremes are properly attenuated, and (c) the resulting audio maintains excellent visual correspondence while achieving noticeably better perceptual quality - sounding more natural and comfortable. This comparison effectively demonstrates how Online-DPO acts as a refinement module that optimizes audio generation quality without compromising content alignment.

\textcolor{blue}{\subsection{Text condition}}

\begin{figure}[t]
\centering
\includegraphics[width=0.95\linewidth]{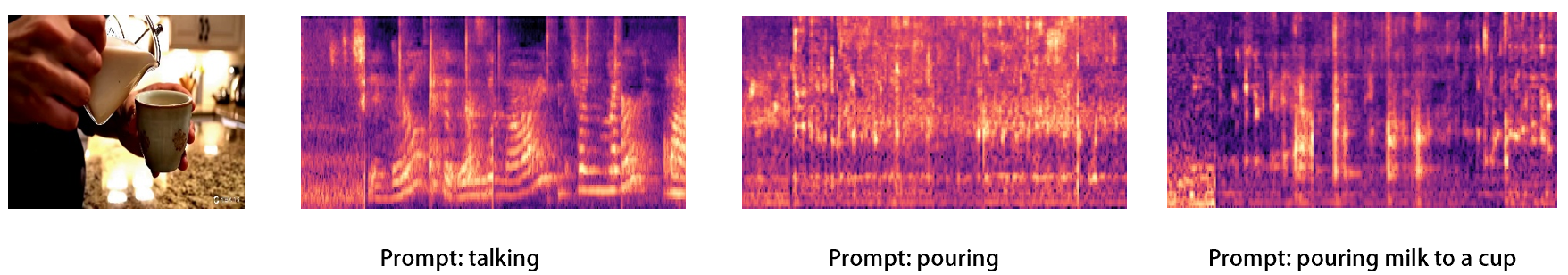}
\caption{\textcolor{blue}{Text conditioning visualization results showing mel spectrograms generated with different text prompts: ``talking", ``pouring", and ``pouring milk to a cup" applied to the same video sequence.}}
\label{fig:vis2}
\end{figure}

\textcolor{blue}{We demonstrate our optional text conditioning capabilities through visualization results in Figure~\ref{fig:vis2}. We evaluate three different text prompts: ``talking", ``pouring" and ``pouring milk to a cup" applied to the same video sequence. }

\textcolor{blue}{When the text prompt is ``talking", the generated audio exhibits characteristic speech patterns in the mel spectrogram, with distinct temporal variations and amplitude fluctuations that correspond to natural speech dynamics. The frequency distribution shows typical vocal formant structures with intermittent energy patterns reflecting conversational rhythm.}

\textcolor{blue}{When the text prompt is ``pouring", the resulting audio demonstrates more continuous and concentrated spectral energy, particularly in higher frequency components. The mel spectrogram reveals sustained, dense frequency content that captures the acoustic characteristics of liquid flow, with consistent energy distribution across time that aligns with pouring actions.}

\textcolor{blue}{Most notably, when using the more specific prompt ``pouring milk to a cup", the generated audio achieves superior alignment with the visual content. The mel spectrogram shows moderated amplitude levels with frequency characteristics that more accurately represent the subtle acoustics of milk being poured into a container. The spectral energy is well-controlled, avoiding excessive loudness while maintaining the essential frequency components associated with liquid-container interaction sounds. This demonstrates that more detailed text descriptions enable our model to generate audio with finer acoustic nuances that better match both the semantic content and the visual context of the scene.}

\subsection{Ablation Study}
\begin{table}[t]
  \centering
  \caption{We evaluated the performance using both CLIP average features and InternVid features for the given video. The InternVid features can produce better performance.}
  \setlength{\tabcolsep}{2pt}
  \begin{tabular}{@{}lccc@{}}
    \toprule
    Method  &MKL $\downarrow$ & CLIP $\uparrow$ & FID $\downarrow$ \\
    \midrule
    CLIP average pooling  &1.727 & 12.01 & 38.93 \\
    Frame features only  &1.842 & 12.95 & 38.35 \\
    InternVid video features  & \textbf{1.379} & \textbf{13.94}& \textbf{32.16} \\
    \botrule
  \end{tabular}
  \label{tab:ab1}
\end{table}
\paragraph{Effect of video condition} 
To evaluate the efficacy of our dual visual feature framework, we conducted comparative experiments with two alternative approaches: (1) replacing the InternVid video encoder with CLIP image encoder using average pooling (following FoleyCrafter's methodology), and (2) utilizing only image frame embeddings. All experiments were performed on the base model (without alignment), with the projection layer and visual-attention mechanism fine-tuned on the VGGSound dataset.

The results in Table~\ref{tab:ab1} reveal several key findings. First, InternVid features achieve superior performance across all metrics, reducing MKL by 6.3\% compared to frame embeddings and by 20.1\% versus CLIP pooling. The CLIP score improvements are equally notable, with +7.6\% and +16.1\% gains respectively. Most significantly, FID scores demonstrate the clearest advantage of video-specific features, showing +4.6\% and +17.4\% improvements. These consistent gains highlight that temporal-aware video features substantially outperform static image-based representations for audio-visual synchronization.

\paragraph{Effect of image frame condition}
\begin{table}[h]
\caption{Comparison of results of image frame condition.}\label{tab2}
\begin{tabular*}{\textwidth}{@{\extracolsep\fill}lccc}
\toprule%
 Method  &MKL $\downarrow$ & CLIP $\uparrow$ & FID $\downarrow$ \\
    \midrule
    Video features only & 1.894& 11.33 & 39.45 \\
    Video + Image frame features  & \textbf{1.379} & \textbf{13.94}& \textbf{32.16} \\
\botrule
\end{tabular*}
\label{tab:ab2}
\end{table}


We conduct a systematic evaluation of our dual visual conditioning approach by comparing two configurations: (1) using only InternVid video-level features, and (2) our complete solution combining both video-level and frame-level features. We show the results in Table~\ref{tab:ab2}.

The combined use of video-level and frame-level features yields a $+0.515$ gain in synchronization accuracy (MKL), demonstrating superior temporal alignment between modalities. For semantic consistency, we observe a +23.0\% improvement in CLIP score, indicating enhanced cross-modal understanding when leveraging both global and local visual information. The audio quality metric shows a +18.5\% enhancement, confirming that hierarchical visual representation leads to more natural sound generation.

 \section{Conclusion}
 In this paper, we present HarmoniDPO, a novel framework leveraging diffusion models and preference optimization for video-to-audio synthesis that generates high-quality audio synchronized with visual input. Our approach uniquely pioneers the application of online Direct Preference Optimization (online-DPO) to the V2A task, enabling the model to learn directly from preferences regarding audio quality and alignment. Our framework utilizes a dual video representation, combining global context with fine-grained frame-wise features, to achieve precise temporal alignment between visual events and generated audio. The integration of online-DPO allows the model to refine outputs based on learned reward functions, significantly improving perceptual quality and video-audio congruence. Additionally, we introduce DDS (Dual-scale Diffusion Search), a novel inference-time optimization algorithm that enhances generation quality through adaptive test-time scaling. Through comprehensive experimental evaluation, we demonstrate that HarmoniDPO consistently outperforms existing approaches in generating high-quality, temporally synchronized audio content that better reflects perceptual preferences and aligns seamlessly with input videos.

\section{Limitation}
\textcolor{blue}{While HarmoniDPO demonstrates significant advancements in video-to-audio synthesis, there are several areas for future improvement. One potential avenue for enhancing performance is the incorporation of higher-quality and larger audio-visual datasets. The largest available dataset, VGGSound, contains fewer than 200,000 samples, with each video limited to just 10 seconds in length. This constraint restricts our model's ability to generalize to longer videos and capture extended temporal dynamics.
Moreover, most existing datasets provide only a single reference audio per video, even though multiple sound interpretations (e.g., different footsteps or background music) can be equally valid. As a result, common evaluation metrics may favor outputs close to the reference while penalizing other plausible generations, leading to a mismatch between automated scores and human perceptual preferences. }




\bmhead{Acknowledgements}
This work was supported in part by the National Key R\&D Program of China (NO.2022ZD0160101).

\bibliography{sn-bibliography}

\end{document}